\documentclass[letterpaper, 10 pt, conference]{ieeeconf}  

\IEEEoverridecommandlockouts                              

\usepackage{graphicx} 
\usepackage{amsmath} 
\usepackage{amssymb}  
\usepackage{multicol}
\usepackage[bookmarks=true]{hyperref}
\usepackage{booktabs} 

\newcommand{\figref}[1]{Fig. \ref{#1}}

\DeclareMathOperator*{\argmin}{arg\,min}
\usepackage{color}

\usepackage[font=footnotesize,labelfont=bf]{caption}
\renewcommand{\baselinestretch}{.91}
\author{Hong J. Jeon and Anca D. Dragan
}

\title{\LARGE \bf
Configuration Space Metrics
}

\begin{document}

\maketitle
\thispagestyle{empty}
\pagestyle{empty}

\begin{abstract}
When robot manipulators decide how to reach for an object, hand it over, or obey some task constraint, they implicitly assume a Euclidean distance metric in their configuration space. Their notion of what makes a configuration closer or further is dictated by this assumption. But different distance metrics will lead to different solutions. What is efficient under a Euclidean metric might not necessarily look the most efficient or natural to a person observing the robot.
In this paper, we analyze the effect of the metric on robot behavior, examining both Euclidean, as well as non-Euclidean metrics -- metrics that make certain joints cheaper, or that correlate different joints. 
Our user data suggests that tasks on a 3DOF arm and the
Jaco 7DOF arm can typically be grouped into ones where a Euclidean metric works well, and tasks where that is no longer the case: there, surprisingly, penalizing elbow motion (and sometimes correlating the shoulder and wrist) leads to solutions that are more aligned with what users prefer.

\end{abstract}

\section{Introduction}

When planning kinematic paths from a start to a goal, robot motion planners often minimize the distance the robot travels in configuration space.
When deciding on a goal configuration within a goal set \cite{dragan_ratliff_srinivasa_2011}, integrating over squared velocities as in trajectory optimization \cite{Ratliff–2009–10204, Schulman:2014:MPS:2675301.2675308}, or summing up the distances between every two consecutive waypoints as in randomized planning \cite{DBLP:journals/corr/abs-1105-1186, DBLP:journals/corr/GammellSB14a, DBLP:journals/corr/JansonP13}, these planners implicitly make an assumption: that the right distance metric to use in configuration space is the \emph{Euclidean} metric. When computing distances between two configurations, they take the difference between the two vectors and compute the squared norm of the difference. 

In this paper, we revisit this implicit assumption. We take inspiration from distance metrics in 
\emph{trajectory space}.
For those spaces, it is established that Euclidean metrics do not work well \cite{Ratliff–2009–10204}: they treat each waypoint along the trajectory as independent. When we want to change one waypoint along a trajectory $\xi$ from $\xi_t$ to a new configuration $q'$, the Euclidean metric would move that single waypoint to $q'$, and keep the rest of the trajectory the same. Call this new trajectory $\xi_I$. This seems intuitive at first, but actually what is special about trajectories is that consecutive waypoints are intrinsically not independent from each other -- they are coupled through \emph{time}.
Prior work has shown how if we correlate consecutive waypoints in a non-Euclidean metric $M$, rather than treating them as independent, we end up with something different from $\xi_I$: we get a $\xi_M$ that not only has the $t^{th}$ waypoint shifted to $q'$, but also smoothly propagates that change to the rest of the trajectory \cite{DBLP:conf/icra/DraganMBS15}. $\xi_M$ is further from $\xi$  according to the Euclidean metric, because more waypoints change. However, according to $M$, $\xi_M$ is actually closer.  

A significant amount of trajectory optimization work utilizes metrics with these temporal correlations.
\cite{Ratliff–2009–10204,Kalakrishnan_RAIIC_2011,Schulman:2014:MPS:2675301.2675308}, which is one way of formalizing the pioneering concept of elastic bands \cite{DBLP:conf/icra/QuinlanK93}.
Prior work has also explored learning the metric from demonstrations of what trajectories \emph{people} find more similar \cite{DBLP:conf/icra/DraganMBS15}.

In this paper, we study whether some of the same ideas
of correlation applied to metrics in the trajectory space also hold for metrics in the configuration space.
In fact, prior work in generating natural motion \cite{gielniak_thomaz_2011} argues for cost functions with potentials between consecutive joints, which can be seen as correlating or anticorrelating consecutive joints in a non-Euclidean configuration space metric. Intuitively, this is akin to the trajectory metric, because joints are coupled too -- not through time, but through the kinematic chain. Similarly, we might think that penalizing movement for different joints differently might be helpful.

\begin{figure}[t!]
    \centering
    \includegraphics[width=\columnwidth]{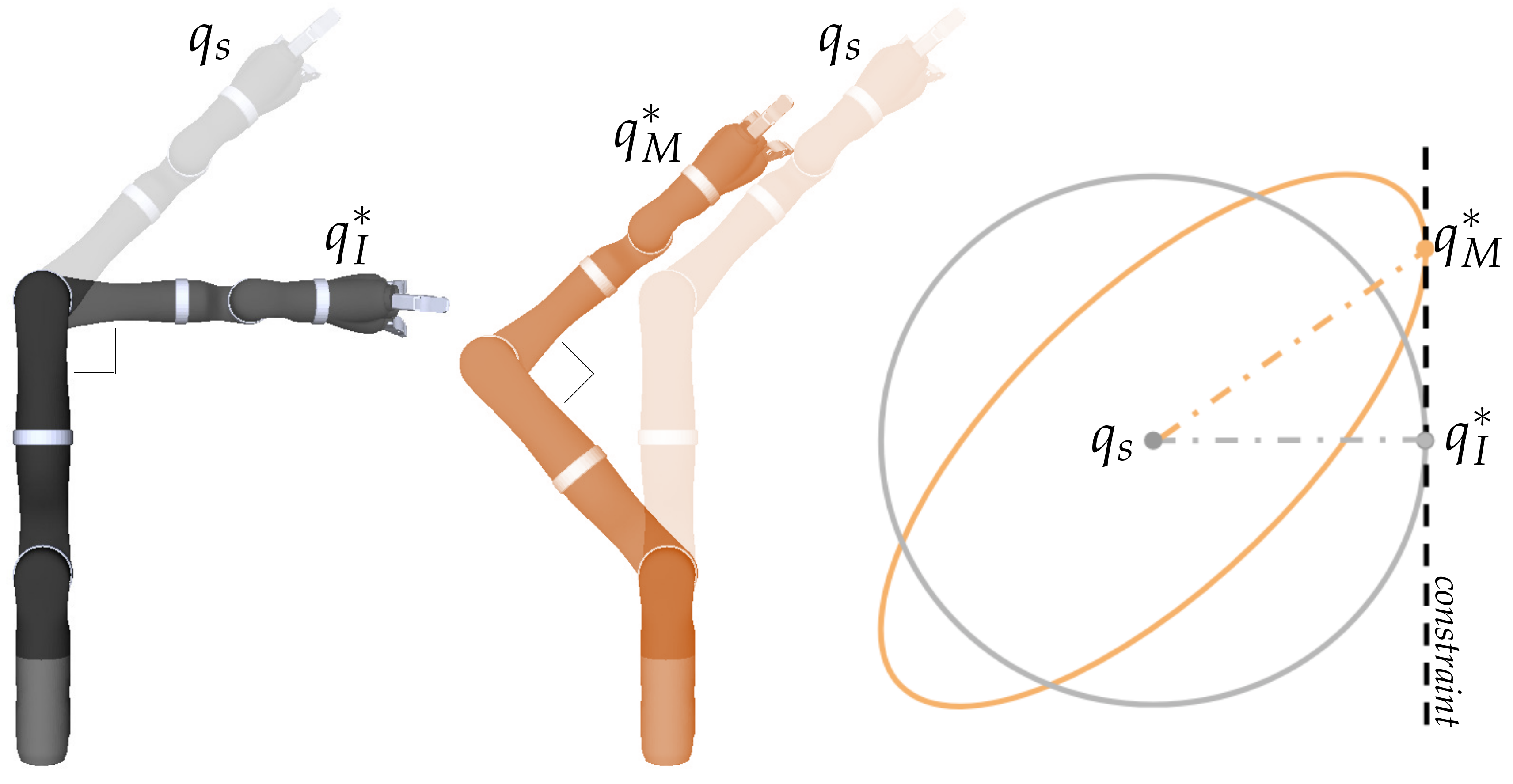}
    \caption{A comparison between the solutions for the Euclidean metric $I$ and a non-Euclidean metric $M$ to the problem of finding the closest configuration to $q_s$ for which the elbow is at $90^\text{o}$. $M$ produces a configuration that looks visually more similar to the $q_s$.}
    \label{fig:cover}
\end{figure}

\figref{fig:cover} illustrates this with an example. The robot starts in an initial configuration $q_s$ and needs to bend the elbow to $90^\textbf{o}$. The Euclidean metric outputs $q_I^*$ as the closest configuration to $q_s$ that satisfies this constraint. But to us, humans,
the configuration on the right, $q_M^*$, actually looks more similar to $q_s$. The Euclidean metric disagrees, but a non-Euclidean metric $M$ warps the space to push $q_I^*$ further away from $q_s$ and bring $q_M^*$ closer (\figref{fig:cover}, right). It does this by coupling the shoulder and the elbow: if the elbow moves to the right, it is cheaper for the shoulder to move to the left to \emph{compensate}, than to stay still.

\begin{figure*}
    \centering
    \includegraphics[height=2.5in, width=6.0in]{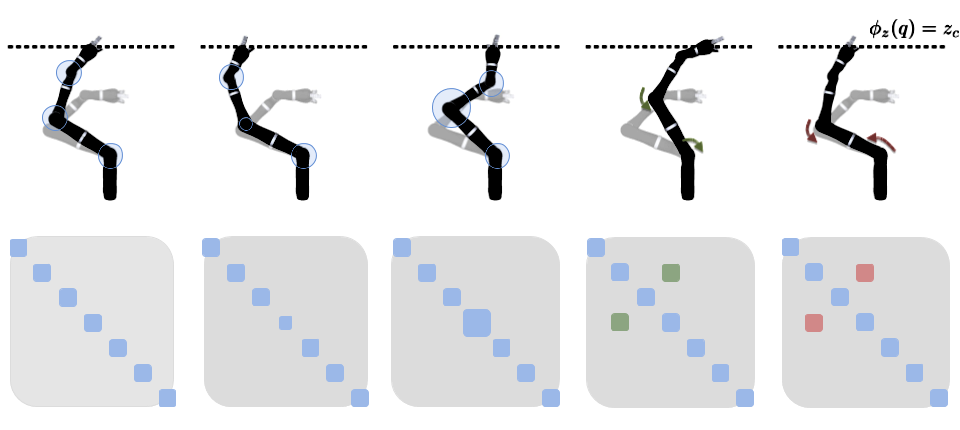}
    \caption{Different metrics lead to visually different solutions for the same task (here, reaching an end effector height).}
    \label{fig:7dof_difference}
    \vspace{-0.2cm}
\end{figure*}

In this paper we make two main contributions towards understanding configuration space metrics:

\noindent\textbf{Understanding the effects of changing the metric.} We start with a 3DOF arm, which enables us to visualize the configuration space, and explore the effects of introducing joint correlations or making joints cheap or expensive. We describe these effects visually, and so characterize ill-conditioned optimization problems that some metrics unfortunately induce. 

\noindent\textbf{Testing whether the Euclidean metric is right.} We then test with both a 3DOF and a 7DOF robot whether the Euclidean metric is the right one with respect to user preferences, under different criteria: producing natural-looking configurations, producing configurations that are visually similar to where the robot starts, and producing configurations that match what people would expect the robot to do. We collect data of user choices, and test how well the Euclidean metric explains these choices, comparing it to a learned metric, designed to best fit the data. 

 Our analysis looks at several tasks that involve varying constraints on the robot's end effector position. We find that tasks 
fall into two groups: 1) Tasks where Euclidean does well, and learning a metric only marginally helps fit real user data and 2) Tasks where the Euclidean metric does poorly. For the latter, metrics that tend to explain user preferences better are similar across criteria, and are rather surprising. They penalize the elbow the most, and not necessarily the shoulder as we might intuit. For 3DOF arms, they correlate the shoulder and the wrist, and not consecutive joints.

Overall, we see evidence that to produce more natural and predictable behavior from robots, we need to change the default understanding of distance in configuration space for certain types of tasks.

\section{Problem Formulation}
We start with an exploration of \emph{whether}, and if so, \emph{how} the choice of the configuration space (C-Space) metric affects what the robot does as a result of an optimization.

The metric defines the robot's understanding of \emph{similarity} or \emph{distance} in its C-Space. Typically, they are optimized to be efficient--- reasoning about what the \emph{shortest} way to achieve the goal is. When we change the definition of distance, we might change what this most efficient solution is.
Goal configurations that were further away may now be closer, and vice-versa, so the robot might choose to approach a given task differently when the notion of efficiency changes.

When faced with a task, like reaching for an object, the robot needs to find a goal configuration that satisfies some constraint $c(q)=0$. For instance, its end effector might need to be at a particular position, or lie on some manifold in task space.
In exploring the effects of different C-Space metrics, we solve constrained optimization problems of the form:
\begin{equation}
\begin{aligned}
q^* =\  & \underset{q \in Q}{\argmin}
& & \|q_s - q\|^2_M  \\
& \text{subject to}
& & c(q) = 0,\\
\end{aligned}
\end{equation}
where $Q$ is the space of robot configurations and $q_s$ is the starting configuration of the robot.
Distance---the notion of closest---is defined via some metric $M$:
\begin{equation}
    \|q_s - q\|^2_M = (q_s - q)^\top M (q_s - q).
\end{equation}
The choice of this metric influence the robot's decision, as we can already see in \figref{fig:7dof_difference} -- we explore this in more detail in the next section.


The constraint here is very important in being able to analyze a metric -- without it, the solution would be $q_s$. Further, problems of this sort appear in motion planning, where we might be interested for instance in moving to the closest configuration that satisfies our task (e.g. a grasping configuration for an object). Which robot configuration is most suitable to plan towards given our current configuration and our constraint?

Formally, a metric is a $d$ by $d$ symmetric and positive definite matrix, where $d$ is the number of degrees of freedom the robot has (also the dimension of $Q$, the C-Space). The metric is a direct result of the inner product we use in the configuration space: 
\begin{equation}
    \langle q_1,q_2 \rangle=q_1^TMq_2
\end{equation}

$M$'s entries can be divided into two groups: diagonal and off-diagonal. These groups lead to two important concepts in understanding the effects of a the metric. Diagonal entries lead to \textit{joint cost}, while the non-diagonal entries lead to \textit{joint correlation}. Next, we will dissect the effects of altering \textit{joint cost} and \textit{joint correlation}.

\begin{figure}
    \centering
    \includegraphics[width=.6\columnwidth]{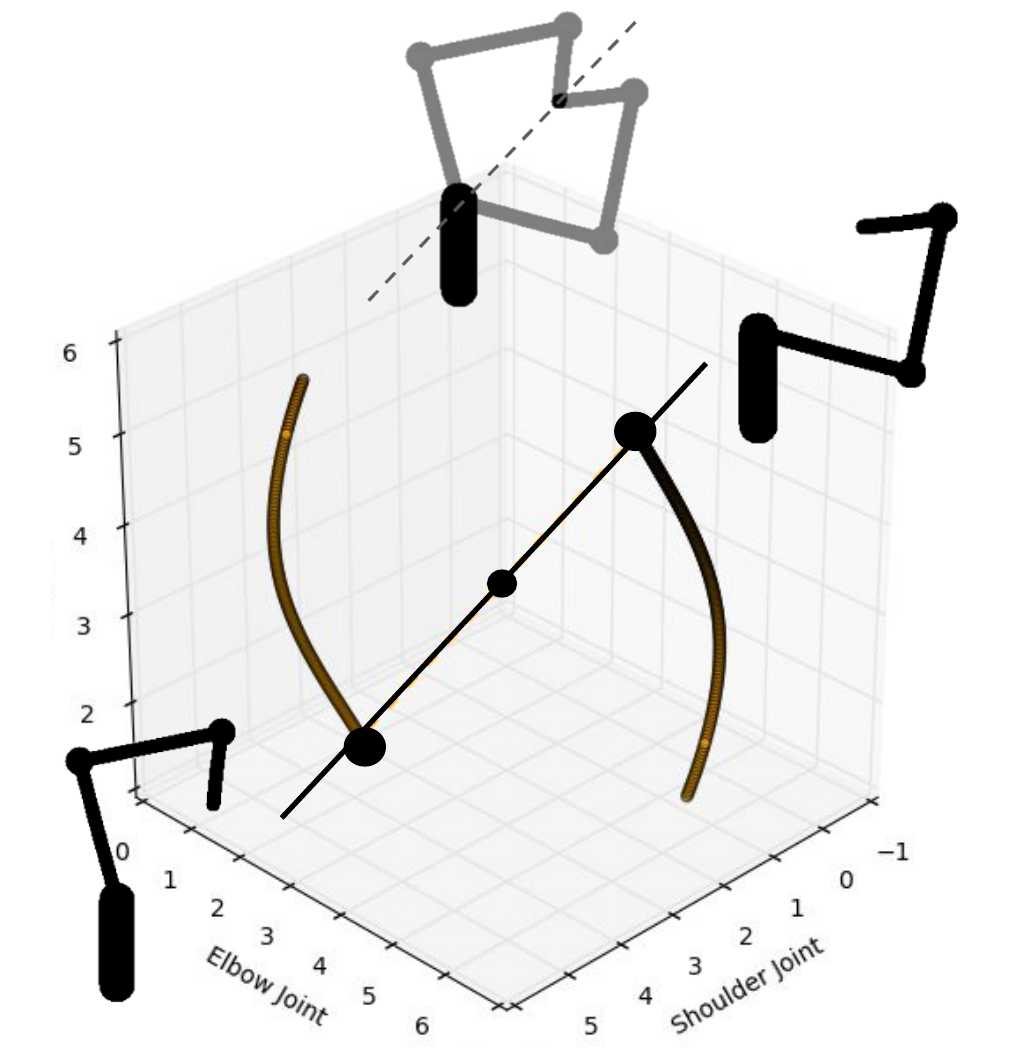}
    \caption{The constraint manifold in C-Space exhibits point symmetry about its centroid. We see that this point symmetry in C-Space translates to reflective symmetry in work space.}
    \label{fig:Symmetry}
\end{figure}

\section{The Effect of the Metric}

We begin with an exploration of robot arms in a simplified, lower dimensional space. Specifically, we consider arms with 3 DOFs operating in a 2D world. The convenience of a 3 DOF arm is that we can visualize its configuration space and gain a better understanding of how the metric affects the robot's choice for how to solve the task. We can plot the feasible set of our constrained optimization problems and \textit{see} how different metrics project onto this set. 

\subsection{Preliminaries}

\begin{figure}
    \centering
    \includegraphics[width=.8\columnwidth]{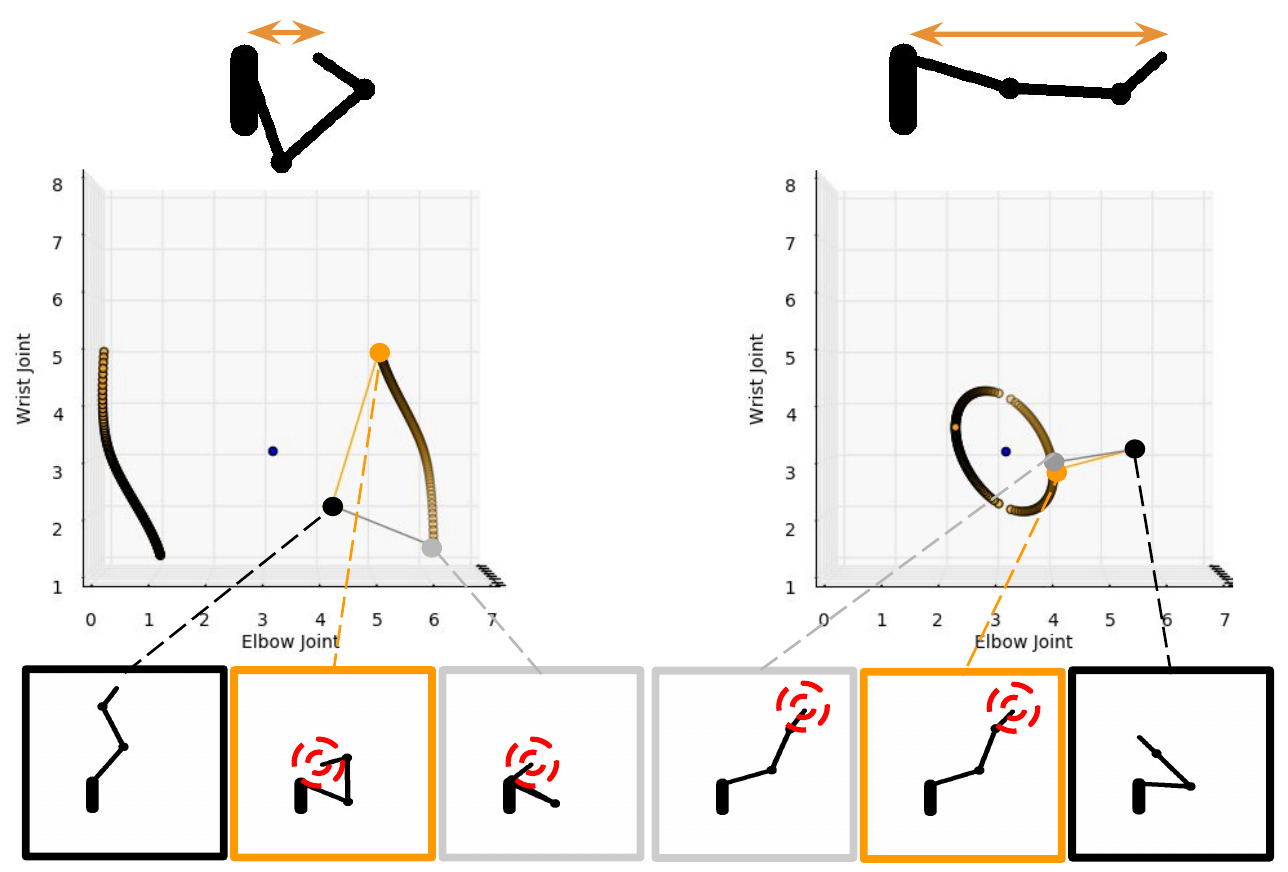}
    \caption{We have 2 sets of start configurations (black border), their closest point on the manifold w.r.t. Euclidean distance (gray border) and an arbitrary different metric (orange). We notice that when we project onto the ``inside" of the manifold (left), we move the end effector closer to the robot base. On the other hand, when projecting onto the ``outside" of the manifold (right), we move the end effector further away.}
    \label{fig:Preliminaries}
    \vspace{-0.2cm}
\end{figure}

\noindent\textbf{End Effector Position Constraints.}
In this work, we will look specifically at constraints on the robot's end effector position.
In \figref{fig:Symmetry} we see the feasible set of such a constraint. It is a 1-manifold with point symmetry with respect to the manifold's centroid. One of the sides corresponds to configurations with elbow less than $\pi$ ("left arm" configurations), while the other corresponds to configurations with the elbow greater than $\pi$ ("right arm" configurations).
\figref{fig:Preliminaries} also shows that as we \emph{increase} the distance between the robot's base and the desired end effector position, the points on the manifold get \emph{closer} to the manifold's centroid. 

\noindent\textbf{Contraction vs. Expansion Tasks.}
 \figref{fig:Preliminaries}(left and right) demonstrates two instances of end effector position constraints. 
The first instance is a \emph{\textbf{contraction}} task. The starting configuration has the end effector further away from the base than the constraint requires. As a result, all the points in the manifold correspond to configurations in which the end effector would ``contract" closer to the base. For such tasks, the starting configuration is on the ``inside" of the manifold. 

The second instance is an \emph{\textbf{expansion}} task. The starting configuration has the end effector closer than it should be, and is on the "outside" of the manifold. 

We notice that in the contraction task, the choice of a metric (Euclidean in gray vs. non-Euclidean in orange) changes the projection point onto the manifold significantly. On the other hand, in the expansion task, the two metrics return nearly identical solutions. Next, we will explore how different metrics influence the optimal solutions, while bearing
in mind that this might be different in expansions vs. contractions.

\subsection{Joint Cost}
Each diagonal term $M_{ii}$ specifies the cost incurred by moving joint $i$: when computing the squared norm of a difference in configurations $q_s-q$, i.e. $\|q_s-q\|^2_M$, $M_{ii}$ weighs the term $(q_s^{(i)}- q^{(i)})^2$---the displacement in joint $i$.

The Euclidean metric weighs all joints equally. However, by breaking this symmetry, we can effectively encourage or discourage the movement of certain joints. The following sections will illustrate that our choice of diagonal weights has intuitive and significant effects on the optimal goal configuration. While it is possible that the Euclidean metric balances the joint costs in just the right way, it is also possible that to achieve our desired goal configurations for the robot, we may want to penalize different joints differently.

\vspace{1em}
\noindent\textbf{Cheap Joints.}
A cheap joint $j$ is one for which $M_{jj} << M_{ii};\quad i \ne j$. When $M_{jj} << M_{ii}$, motion along joint $j$ incurs negligible cost relative to motion along other joints. This has a simple effect: the robot moves joint $j$ more in order to spare motion in the other joints. As a result, when minimizing $\|q_s-q\|_M^2$ for a cheap joint metric $M$, we reduce this $3$ dimensional norm minimization to a $2$ dimensional minimization because cost in the cheap joint is negligible.

\begin{figure*}
    \centering
    \includegraphics[height=3.0in, width=7.0in]{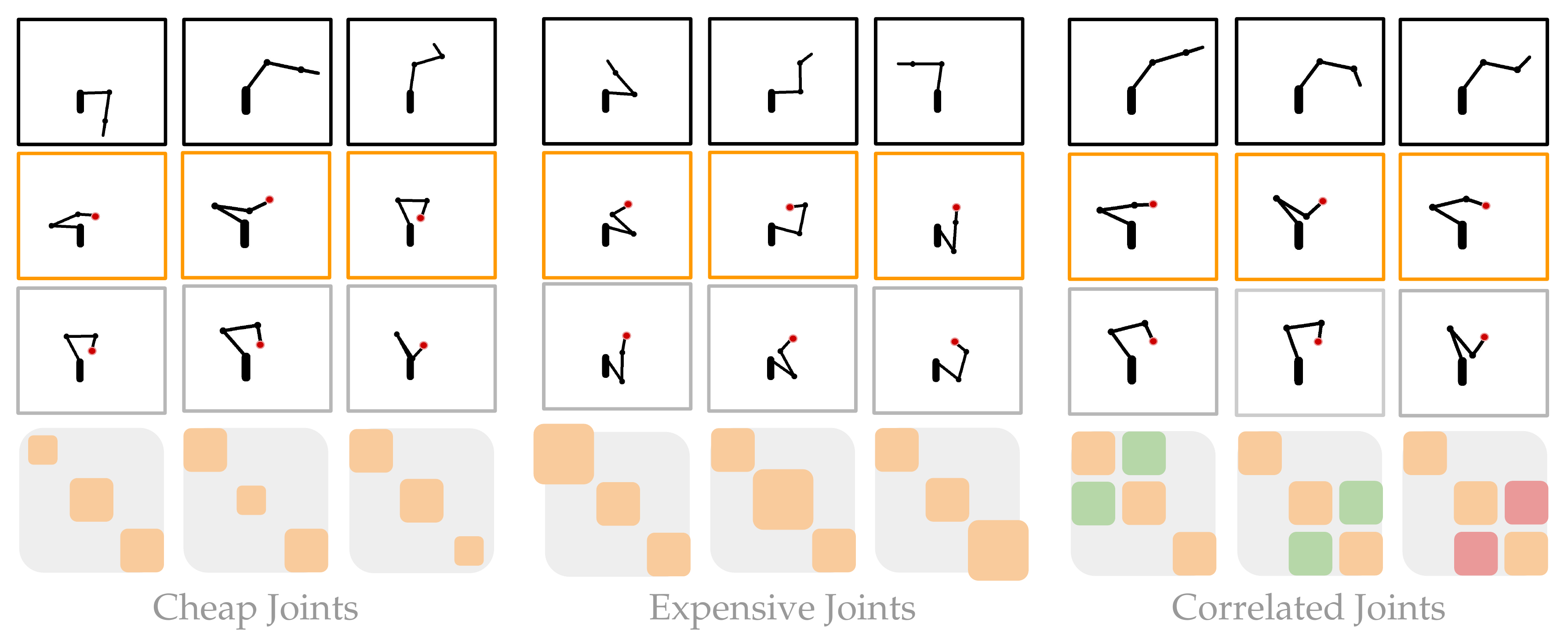}
    \caption{Intuitive effects of different metrics on the solutions to end effector location constraints (red dots). Black border is the start configuration. Orange border is the solution of the metric shown on the bottom row. Gray border is the Euclidean metric's solution. Green tiles denote positive correlation and red ones negative.}
    \label{fig:Intuitive}
    \vspace{-0.2cm}
\end{figure*}
\begin{figure}
    \centering
    \includegraphics[width=\columnwidth]{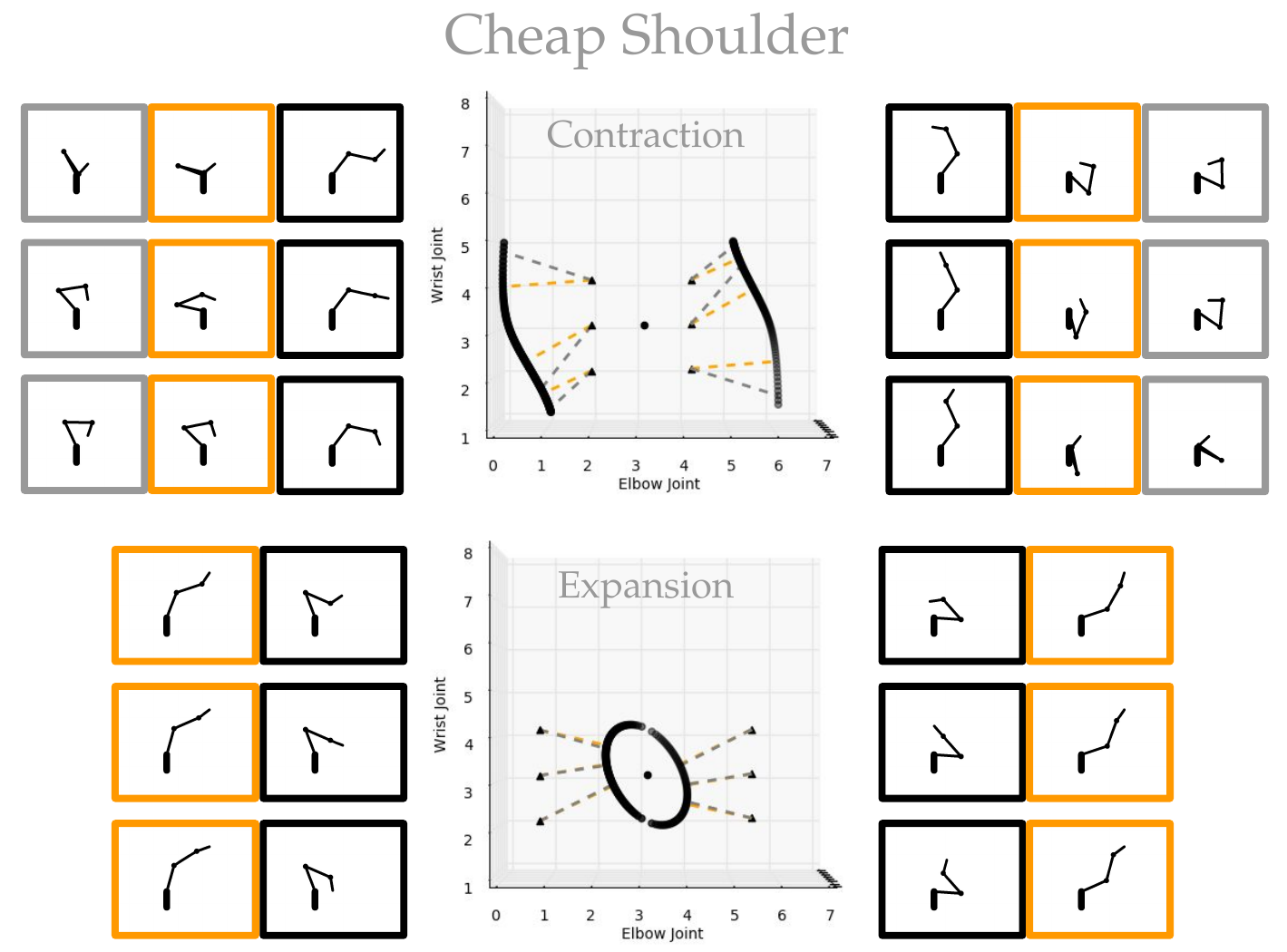}
    \caption{The effects of a cheap shoulder metric on contraction tasks (top). While this metric induces more shoulder movement, it also as a result involves less wrist movement. Meanwhile, for expansion tasks, we see no distinction between the Euclidean metric's solutions and the cheap shoulder's (bottom).}
    \label{fig:CheapShoulder}
    \vspace{-0.2cm}
\end{figure}

\vspace{1em}
\noindent\textit{Cheap Shoulder:}  In \figref{fig:Intuitive}, the first column shows the effect of a cheap shoulder metric. This metric (solution in orange) moves the shoulder significantly more than the Euclidean metric (gray) relative to the starting configuration (black).
\figref{fig:CheapShoulder}(top) shows that this is in general reflected for many start configurations in contraction tasks. On the other hand, all expansion tasks tended to produce very similar results with the two metrics, as we saw before in the preliminaries. 

\vspace{1em}
\noindent\textit{Cheap Elbow:} The second column in \figref{fig:Intuitive} shows an example comparing a cheap elbow metric and the Euclidean one. Again, the elbow moves more to reduce motion in the wrist and shoulder. For this metric too, our analysis revealed that expansion tasks led to smaller differences. For contraction tasks differences were only large when the wrist was close to $\pi$, as in the example from \figref{fig:Intuitive}. 

\figref{fig:7dof_difference} (second column) shows a cheap elbow metric on a 7DOF arm (where the constraint is reaching an end effector height). Compared to Euclidean (first column), we again see that the robot moves its elbow more in order to reduce movement in the shoulder and the wrist.



\begin{figure}
    \centering
    \includegraphics[width=.75\columnwidth]{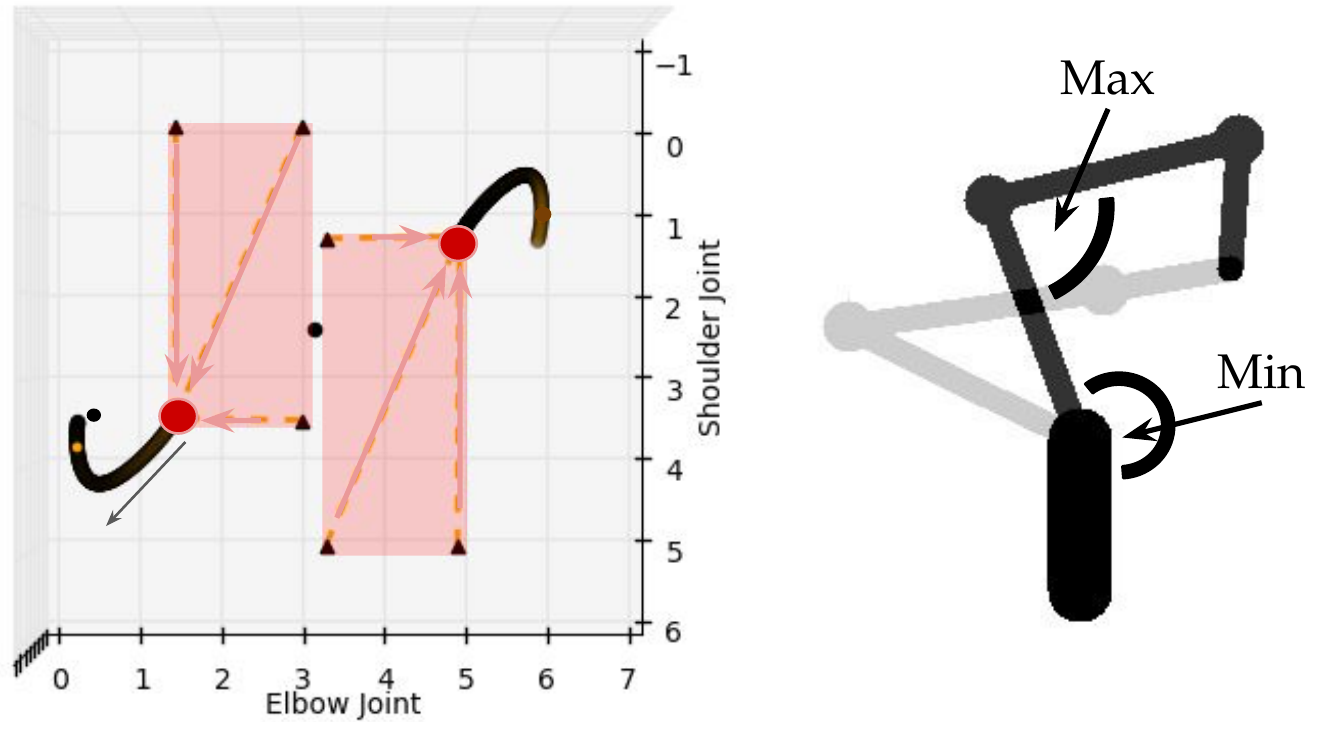}
    \caption{(left) we see that all starting configurations in the red volumes map to their unique solution (Red dot). (Note volume, not area because of the wrist dimension). In the left red volume, this is because the point on the manifold with maximum elbow value coincides with the point of minimum shoulder value. Vice versa for the right red volume.}
    \label{fig:Problems-singluarity}
\end{figure}

\vspace{1em}
\noindent\textit{Cheap Wrist:} 
A cheap wrist solution is in \figref{fig:Intuitive}. With this metric, the shape of the manifold is such that many starting points end up being projected to the same two configurations (\figref{fig:Problems-singluarity}, left). Looking at only the left side of the manifold, all configurations in the red shaded region project to the configuration corresponding to the maximum elbow joint value and the minimum shoulder joint value. Such a point exists with cheap wrist metrics because they create manifolds in which the point with the maximum elbow joint value coincides with the point with minimum shoulder joint value. \figref{fig:Problems-singluarity}(right) depicts why: the solid arm is the configuration with the maximum elbow value and reducing the elbow value only increases (and can not decrease) the shoulder value, thus corresponding to the minimum shoulder value as well.

\vspace{1em}
\noindent\textbf{Expensive Joints}
A joint $j$ is expensive when $M_{jj} >> M_{ii};\ i\ne j$. In this case, the robot moves joint $j$ as little as possible. While cheap joints reduced a $3D$ distance minimization to a $2D$ minimization, expensive joints reduce $3D$ to $1D$.

\vspace{1em}
\noindent\textit{Expensive Shoulder:} In \figref{fig:Intuitive} (4th column) we see an intuitive instance of the expensive shoulder metric. The Euclidean metric moves the shoulder a considerable amount to reach the end effector location, but the expensive shoulder metric barely moves it at all. 

\begin{figure}
    \centering
    \includegraphics[width=1.4in]{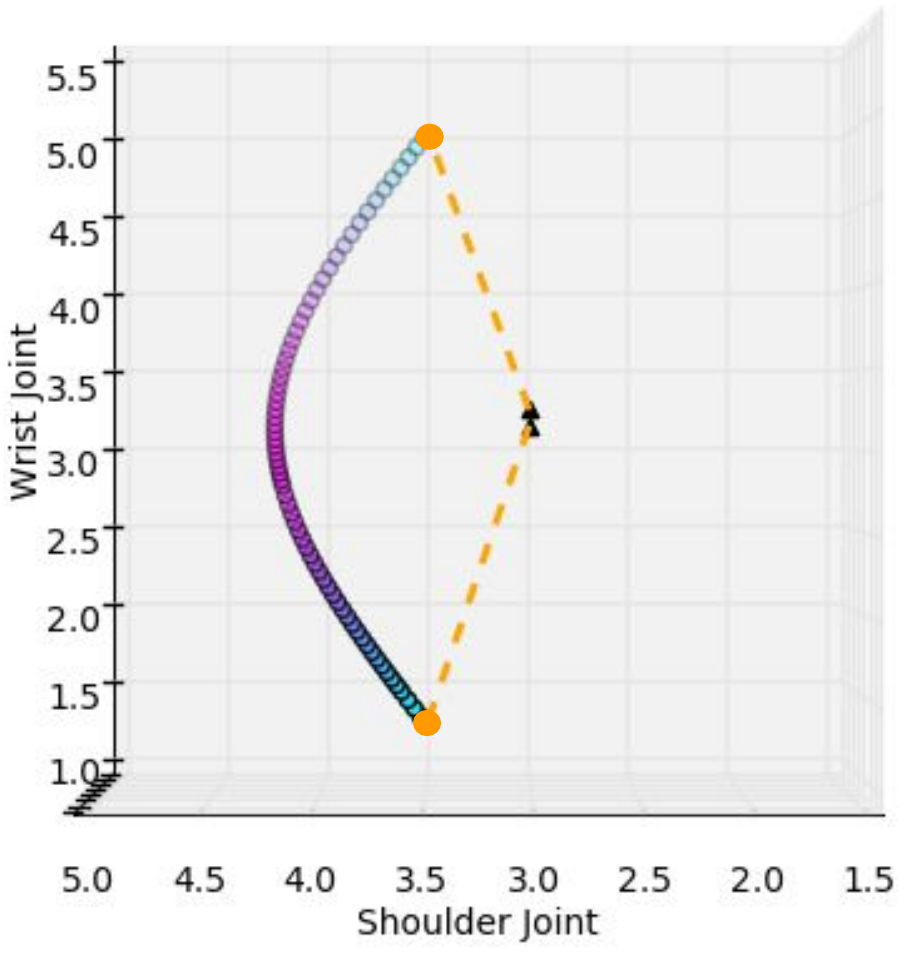}
    \caption{The above metric induces ill-conditioning. The optimal cyan regions are disjoint, causing small changes in the start configuration (black dots) to drastically change the solutions (orange dots) on the manifold}
    \label{fig:Problems-instability}
    \vspace{-0.2cm}
\end{figure}

For contraction tasks, when we minimize just over this one dimension, we unfortunately experience ill-conditioned behavior.
We define \textit{ill-conditioning} as a scenario when the lowest cost sub-level set defined by the metric for a given starting configuration is disjoint. \figref{fig:Problems-instability} illustrates this phenomenon by encoding distance to the manifold in a heat-map (cyan meaning close, lavender meaning far). The cyan regions are at completely opposite ends of the manifold. This behavior occurs because we are minimizing distance in the shoulder dimension while our manifold has reflective symmetry across the shoulder joint, always giving us 2 equidistant solutions.

As a result, two nearly identical starting configurations map to two very distinct solutions.
Note that ill-conditioning only appears in contraction tasks, and not in expansion tasks.

\vspace{1em}
\noindent\textit{Expensive Elbow:} Expensive elbow metrics minimize movement in the elbow. \figref{fig:Intuitive} (5th column) gives a simple example. \figref{fig:7dof_difference} (3rd column) shows the same for the 7DOF robot -- compared to Euclidean, the elbow barely moves, whereas the shoulder and especially wrist move a lot more. 

However, in general, expensive elbow metrics leads to a singularity in the optimization in both contraction and expansion tasks for the 3DOF robot. We notice that an expensive elbow metric behaves identically to a cheap wrist metric for these tasks. For contraction tasks this is intuitive: to contract an arm, the robot must reel in its elbow joint. There must exist a unique configuration that minimizes the amount we adjust the elbow while still allowing the robot's end effector to reach the desired location. This one configuration is shown in red in \figref{fig:Problems-singluarity}. The same follows for expansion tasks. 

\vspace{1em}
\noindent\textit{Expensive Wrist:} Expensive wrist metrics are adverse to moving the wrist. \figref{fig:Intuitive}(6th column) demonstrates such a scenario. In general, such a metric is more well-behaved, in the sense that different starting configurations project to different solutions if they have a different wrist value.


\subsection{Joint Correlation}
The off-diagonal term $M_{ij}$ specifies the correlation between joints $i$ and $j$, because it weighs the term $\Delta q_i\Delta q_j$ (and $\Delta q_j\Delta q_i$) where $\Delta q_i = q_s^{(i)}-q^{(i)}$. 

The Euclidean metric has no correlation between joints. However, by applying correlations, we can encourage certain joints to move together. If $M_{ij}$ is negative, the robot is incentivized to move joints $i$ and $j$ together in the same direction. On the other hand, if $M_{ij}$ is positive, the robot prefers moving joints $i$ and $j$ together in opposite directions. From a biological standpoint, human arms exhibit a degree of coupling in muscles/joints, so we might expect that natural metrics might be non-Euclidean \cite{10.1007/978-3-642-02809-0_9}, \cite{10.1371/journal.pone.0164050}.

\begin{figure}
    \centering
    \includegraphics[width=\columnwidth]{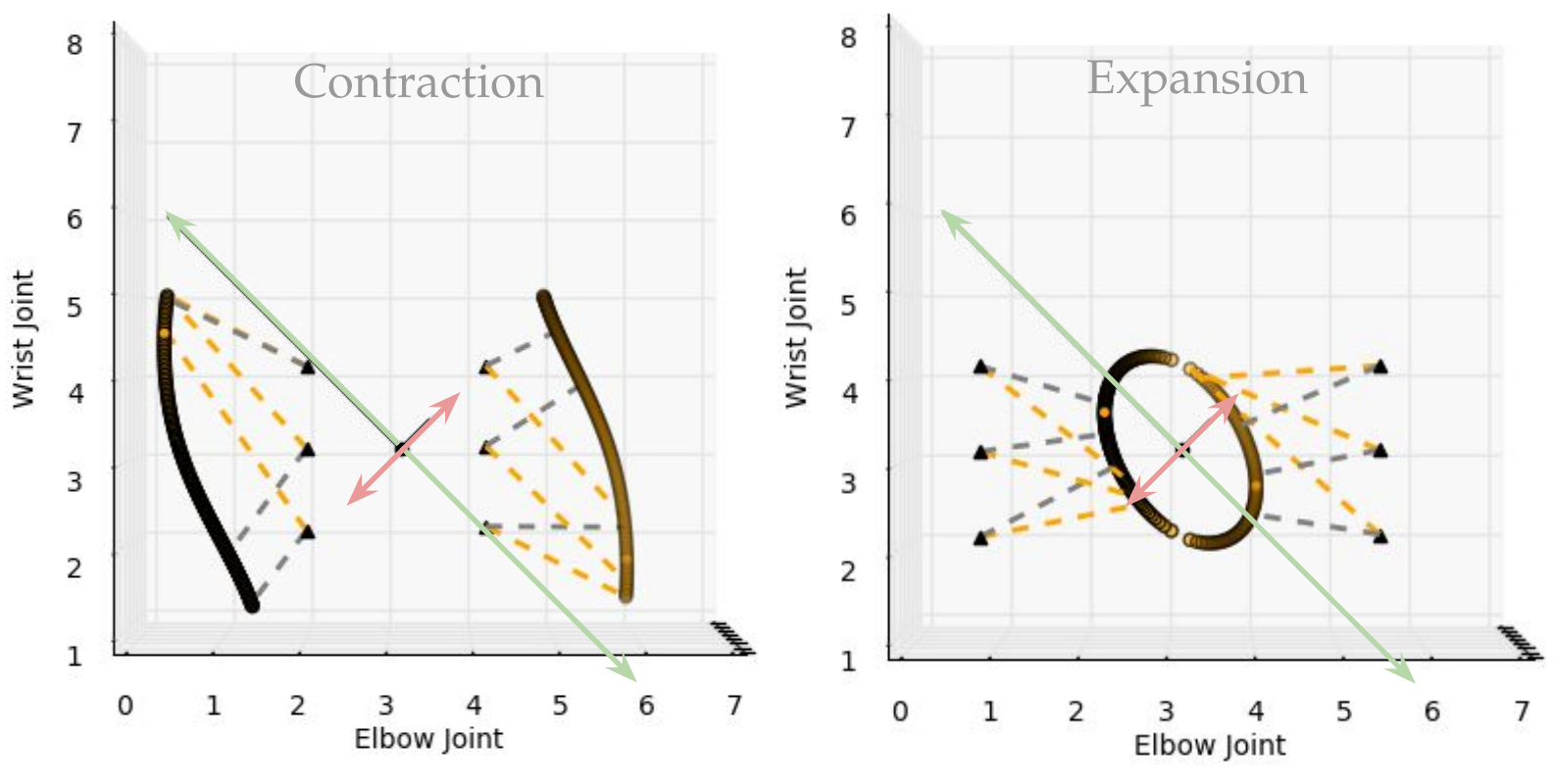}
    \caption{Joint correlation creates directions of low cost in C-Space. The green arrows denote this low cost direction and as a result, the projections run $\sim$ parallel to the green arrows.}
    \label{fig:Correlation}
    \vspace{-0.2cm}
\end{figure}

\figref{fig:Intuitive} (columns 7 through 9) show examples of correlations: positive shoulder-elbow, positive elbow-wrist, and negative elbow-wrist. In each situation, the joints move together in the opposite or in the same directions, while with Euclidean they move independently. \figref{fig:7dof_difference} also shows a positive (4th column) and a negative (last column) correlation between shoulder yaw and elbow. Compared to Euclidean, the positive correlation ends up moving the shoulder clockwise, because the elbow has moved counterclockwise. This is a prime example where there is clearly more total motion than Euclidean (by definition), and yet the resulting configuration would believably look much more natural to some users. 

\figref{fig:Correlation} depicts the projections onto the manifold for correlations. Correlations induce an axis along which movement is cheap, and projections (orange) tend to run parallel to this axis compared to Euclidean (gray).

\section{Is Euclidean the Right Metric?}
In the previous section, we saw that different metrics can lead to different outcomes, especially for contraction tasks. This begs the question of which metric a robot should be optimized over. The answer depends on what our objective is. If we want spatial efficiency, the Euclidean metric is an obvious choice. But what if we want the robot to move more naturally? Or predictably?

We investigate the fit of the Euclidean metric as the answer to such questions. We do so by leveraging learning as a tool: we ask people which configurations they find more natural, visually similar to the starting configuration, predictable, etc. We then learn a metric that agrees with their answers, analyze its characteristics,
and compare it to the Euclidean metric. If the Euclidean metric agrees with user answers almost as well as the learned metric, that suggests that Euclidean is actually a good choice. Otherwise, we may need to reconsider our notion of efficiency when operating around people.


\subsection{Learning a Metric}
Now that we have a general intuition of C-Space metrics, we can consider learning metrics for different criteria (for example: naturalness, predictability or visually similarity). 

To learn a metric, we use preference-based learning: users provide their preferred solutions among several alternatives, and use their answer as evidence about the metric they are using to determine the solution. 

We chose preference-based learning for our analysis because it is actually feasible to collect preferences from end users. In contrast, asking users to demonstrate the solution is not only more burdensome, but configuration spaces are counterintuitive enough to make figuring out the desired configuration difficult to impossible. 



With preference-based learning, our framework consists of $n$ multiple choice questions $Q_1, Q_2, ..., Q_n$. Each question consists of a robot in its starting configuration $q^{(s)}_{i}$ followed by $m$ feasible goal configurations that make up the answer choices: $Q_i = \{q^{(s)}_{i}, q^{(1)}_{i}, q^{(2)}_{i}, ..., q^{(m)}_{i}\}$. 

We ask users to select the answer choice that best fits a given objective, like naturalness. 
Of course, not every person will select the same answer.  When we aggregate responses across multiple users, we produce a distribution $f(Q_i)$ over answer choices for each question. This distribution reflects both which solutions users preferred more and also by how much they preferred it to others. The metric should produce shorter distances to more popular solutions. Furthermore, the difference in distance should be larger if the subjects greatly favor one solution over another.

With this in mind, we fit a metric that minimizes the Kullback$-$Leibler Divergence between the distribution induced by user answers and the one induced by the metric:
\begin{equation}
\begin{aligned}
M^* =\  & \underset{M \in S^{++}}{\argmin}
& \sum_{i=1}^{n} D_{KL}[f(Q_i) | \sigma(M, q^{(s)}_i)] \\
\end{aligned}
\end{equation}
where 
\begin{equation}
    \sigma(M, q^{(s)}_i)_j = \frac{e^{-\|q^{(s)}_{i} - q^{(j)}_i\|_M^2}}{\sum_{j=1}^{m} e^{-\|q^{(s)}_{i} - q^{(j)}_{i}\|_M^2}}
\end{equation}
the softmax negative squared distances defined by metric $M$ \cite{conf/aaai/ZiebartMBD08}, \cite{bradley_terry_1952}, \cite{luce_2005}. 
With this optimization, we learn a metric that attributes low cost to more frequently selected answer choices while keeping answer choices of similar frequency at the same distance.

To solve this optimization, we perform gradient descent on a matrix variable $M$. Since softmax is sensitive to scaling of $M$, we apply the constraint $\|M\|_F = 1.0$ on top of positive definiteness. The matrix variable $M$ can be thought of as $6$ scalar variables $M_{11}, M_{22}, M_{33}, M_{12}, M_{13}, M_{23}$ which represent the entries of matrix $M$. The quadratic form $\|q^{(s)}-q^{(j)}\|_{M}$ is therefore linear in these variables $M_{ab}$. The negative softmax KL Divergence results in:
\begin{equation}
    \begin{aligned}
        &\sum_{i=1}^{n} \sum_{j=1}^{m}& f(Q_i)_j \bigg[\log \Big(\frac{f(Q_i)_j*\sum_{k=1}^{m} e^{-\|q^{(s)}_{i} - q^{(k)}_{i}\|_M^2}}{e^{-\|q^{(s)}_{i} - q^{(j)}_i\|_M^2}}\Big)\bigg]\\
        = &\sum_{i=1}^{n} \sum_{j=1}^{m}& f(Q_i)_j*\big[LSE(d^{(1)}_{i}, ..., d^{(m)}_{i}) + \|q^{(s)}_{i} - q^{(j)}_i\|_M^2\big]\\
    \end{aligned}
\end{equation}

\begin{table}
    \vspace{0.4cm}
    \caption{KL Divergence for Euclidean and Learned Metric}
    \label{tab:freq}
    \centering
    \begin{tabular}{ccc}
    \toprule
    & Euclidean & Learned\\
    \toprule
    Naturalness Contraction & 8.531006144 & 2.090473224 \\
    Naturalness Expansion & 3.840278556 & 1.900694689 \\
    \midrule
    Similarity Contraction & 12.0261777 & 2.520342637 \\
    Similarity Expansion & 2.666917419 & 1.664360558 \\
    \midrule
    Closeness Contraction & 13.22421051 & 2.141058327 \\
    Closeness Expansion & 2.699116968 & 1.968807375 \\
    \midrule
    Predictability Contraction & 6.767907939 & 1.594664342 \\
    Predictability Expansion & 2.320071361 & 1.66843745 \\
  \bottomrule
\end{tabular}\label{tab:kl}
\vspace{-0.4cm}
\end{table}

This is a non-negative linear combination of convex functions so the overall objective is convex as well.

\subsection{Data}
\noindent\textbf{Queries.} We systematically generated the multiple choice queries. Firstly, we disallowed the elbow joint to ``flip" (cross $q[2] = \pi$) while traveling from start to goal. 
Afterwards, we applied wrist joint limits and eliminated self collisions. Finally, we sorted the configurations by increasing wrist value and picked $m$ points uniformly across this sorted set. The uniform selection gives us high chances of providing at least one good solution while keeping each the solutions different enough from each other.

Each multiple choice question had $1$ image of a robot starting configuration along with $4$ images as answer choices generated from the strategy described above. Along with the robot, each image had a red dot specifying the location that the robot end effector must reach.

There were 36 questions (18 contractions, 18 expansions).

\noindent\textbf{Subjects. } Each participant answered all 36 multiple choice questions. We recruited $23$ participants (mean age of 34, female 48$\%$) via Amazon Mechanical Turk. All participants were from the United States and had a minimal approval rating of $95\%$.

\noindent\textbf{Criteria for Metrics.} We were interested in learning metrics for various human preferences so we decided to ask for 4 separate answers for each query.
\begin{enumerate}
    \item \emph{\textbf{Naturalness: }}In which answer choice does the robot look most \textbf{natural}?
    \item \emph{\textbf{Visual similarity: }}In which answer choice does the robot look most \textbf{visually similar} to the start position?
    \item \emph{\textbf{Closeness: }}In which answer choice does is the robot \textbf{closest} to the start position?
    \item \emph{\textbf{Predictability: }}In which answer choice does the robot move how you would \textbf{expect} given the start configuration and red dot?
\end{enumerate}
Each participant answered all $4$ sub parts of each question and we used their responses to optimize $4$ separate metrics.  

\subsection{Analysis} 
With the user data, we learned an expansion and contraction metric for each of the 4 criteria. 

\vspace{1em}
\noindent\textbf{Euclidean better for expansion.} Table \ref{tab:kl} compares the KL Divergence of the learned metrics with those of the Euclidean metric.
We first notice that across all 4 criteria, the Euclidean metric's KL Divergence is significantly lower for expansion tasks than contraction ones. This suggests that the Euclidean metric is a better fit for expansion tasks, and that perhaps other metrics might be more suitable for contraction tasks. 

\begin{figure}
    \centering
    \includegraphics[width=.8\columnwidth]{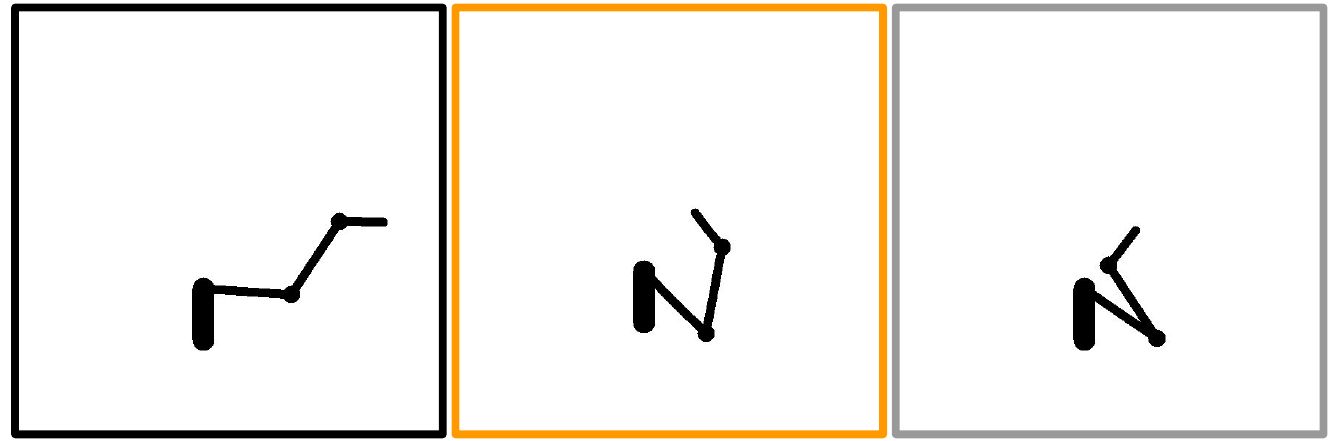}
    \caption{The effects of our learned contraction metric (orange). Compared to the Euclidean metric (gray), the learned metric produces a much more natural looking solution. Participants also found this configuration more visually similar, close, and predictable.
    }
    \label{fig:LearnResults}
    \vspace{-0.2cm}
\end{figure}

This is in line with our analysis from the previous section. Firstly, expansion tasks have a smaller feasible solution set than contraction tasks. There are less ways to reach for a far away object than to reach for a close one. In order to reach far away objects, our wrist and elbow joint angles must be to some degree straight. We can observe this by comparing the solution sets of expansion and contraction tasks in \figref{fig:CheapShoulder}. The lower subfigure displays an expansion task's solution set and it 1) is more compact than its contraction counterpart 2) avoids joint limits in all $3$ joints. Solutions near joint limits are almost always unnatural looking. As a result they are likely not the solutions that users would predict the robot to move to or consider "visually similar" to more natural looking start configurations.

\vspace{1em}
\noindent\textbf{Learned metrics for contraction.} Across all 4 criteria, the metrics learned for contraction tasks were nearly indistinguishable, and all reached much lower KL Divergence than the Euclidean. 
Key features of these metrics were:
\begin{itemize}
    \item expensive elbow joint,
    \item very strong ($\sim$.99) positive correlation of shoulder and wrist,
    \item moderate ($\sim$.50) positive correlation of shoulder and elbow, and moderate ($\sim$.50) positive correlation of the elbow and wrist.
\end{itemize} \figref{fig:LearnResults} illustrates the effects of this metric. We notice the significant effect of shoulder/wrist positive correlation. The orange bordered configuration looks
especially more natural. The Euclidean metric in comparison maps to an uncomfortable contracted positions that users probably disliked.

\vspace{1em}
\noindent\textbf{Learned metrics for expansion.} Across all 4 criteria, the learned metrics were different from Euclidean, but only fit the user data slightly better. Interestingly, different criteria led to different metrics.

\vspace{1em}
\noindent\textit{Expansion Naturalness Learned Metric:}
The learned naturalness metric had:
\begin{itemize}
    \item small amount ($\sim$.3) of positive correlation along shoulder and elbow.
    \item small amount ($\sim$.3) of positive correlation along elbow and wrist.
    \item small amount ($\sim$-.25) of negative correlation along shoulder and wrist. 
    \item expensive elbow, neutral shoulder, cheap wrist.
\end{itemize}
The magnitude of work space change is noticeably smaller for expansion tasks than contraction ones. This agrees with the observation that Euclidean metric KL Divergence was significantly lower for expansion tasks. However, among the notable differences was a concerted effort to keep the wrist angle near $\pi$ while not in a singularity. This is understandable for a naturalness metric because the wrist is often perfectly straight when reaching for distant locations. The absence of a singularity suggests that while users want the wrist to be straight, they prefer that their goal configuration, to a degree, resembles their start configuration. 

\vspace{1em}
\noindent\textit{Expansion Visual Similarity and Closeness.}
Visual Similarity and Closeness learning converged to nearly identical metrics. Some notable characteristics were
\begin{itemize}
    \item strong ($\sim$.97) positive correlation between the shoulder and elbow,
    \item  moderate ($\sim$.4) positive correlation between the shoulder and wrist,
    \item negligible correlation in the elbow and wrist,
    \item and a cheap shoulder (by an order of magnitude).
\end{itemize}   

\vspace{1em}
\noindent\textit{Expansion Predictability:}
The predictability metric had
\begin{itemize}
    \item moderate ($\sim$0.70) positive correlation in the shoulder and elbow,
    \item  moderate ($\sim$ 0.65) negative correlation in the shoulder and wrist, 
    \item expensive elbow, neutral shoulder, cheap wrist
\end{itemize}
The Euclidean had larger spread and the learned metric resisted spread, generating solutions with wrist value $\sim \pi$.

\vspace{1em}
\noindent\textbf{Summary.} Overall, the Euclidean metric did not seem like the best fit for contraction tasks, where the learned metric consistently ended up with an expensive elbow and a strong correlation of the shoulder and wrist. 

We had expected the shoulder to be most expensive, since it is higher up the kinematic chain (instead it was the elbow), and for consecutive joints to correlate (instead they only had moderate correlations, the strongest being shoulder-wrist). 

We find this important, because it teaches us not only that we can't necessarily rely on the Euclidean metric as a default, but also that users might contradict our intuition.

\section{Metrics in 7DOFs for the Jaco 7}
\begin{figure}
    \centering
    \includegraphics[width=0.9\columnwidth]{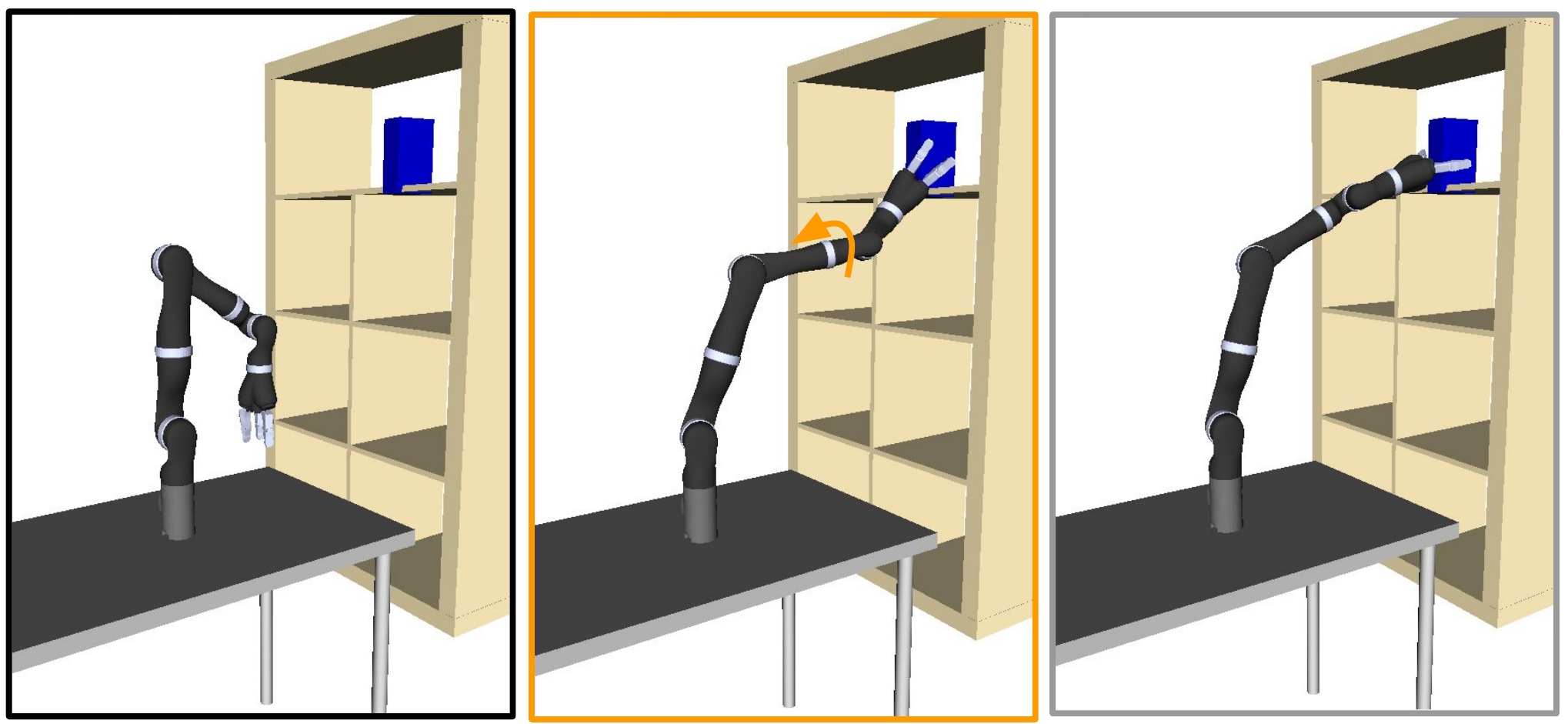}
    \caption{The robot is situated at a starting configuration (black). It must reach the book in the shelf. Notice that with the learned metric (orange), the robot learned to rotate the wrist roll joint to reach the book. This is a motion we as humans very liberally take. The Euclidean metric (gray) makes no use of these auxiliary joints and as a result, is resigned to stiff, robotic solutions.}
    \label{fig:Learned7DOF}
\end{figure}

Upscaling to 7DOF arms in 3D task space, we can learn metrics that encode even more.
\figref{fig:Learned7DOF} illustrates a fascinating behavior that a non-Euclidean metric learned. Instead of constraining its motion to a plane and actuating shoulder pitch, elbow, and wrist pitch to reach the book in the shelf, the metric (orange) learned to use the wrist roll to rotate the wrist and reach the book. As humans, we use this motion all the time when picking up or reaching for objects (Imagine picking fruit from a tree. While your hand may start down by your side, you will flip the orientation of your wrist/forearm to eventually grab the fruit with your palm facing you.) This human-like motion is rarely replicated via a Euclidean metric (gray).

\begin{figure}
    \centering
    \begin{tabular}{cc}
        \includegraphics[height=1.in, width=1.in]{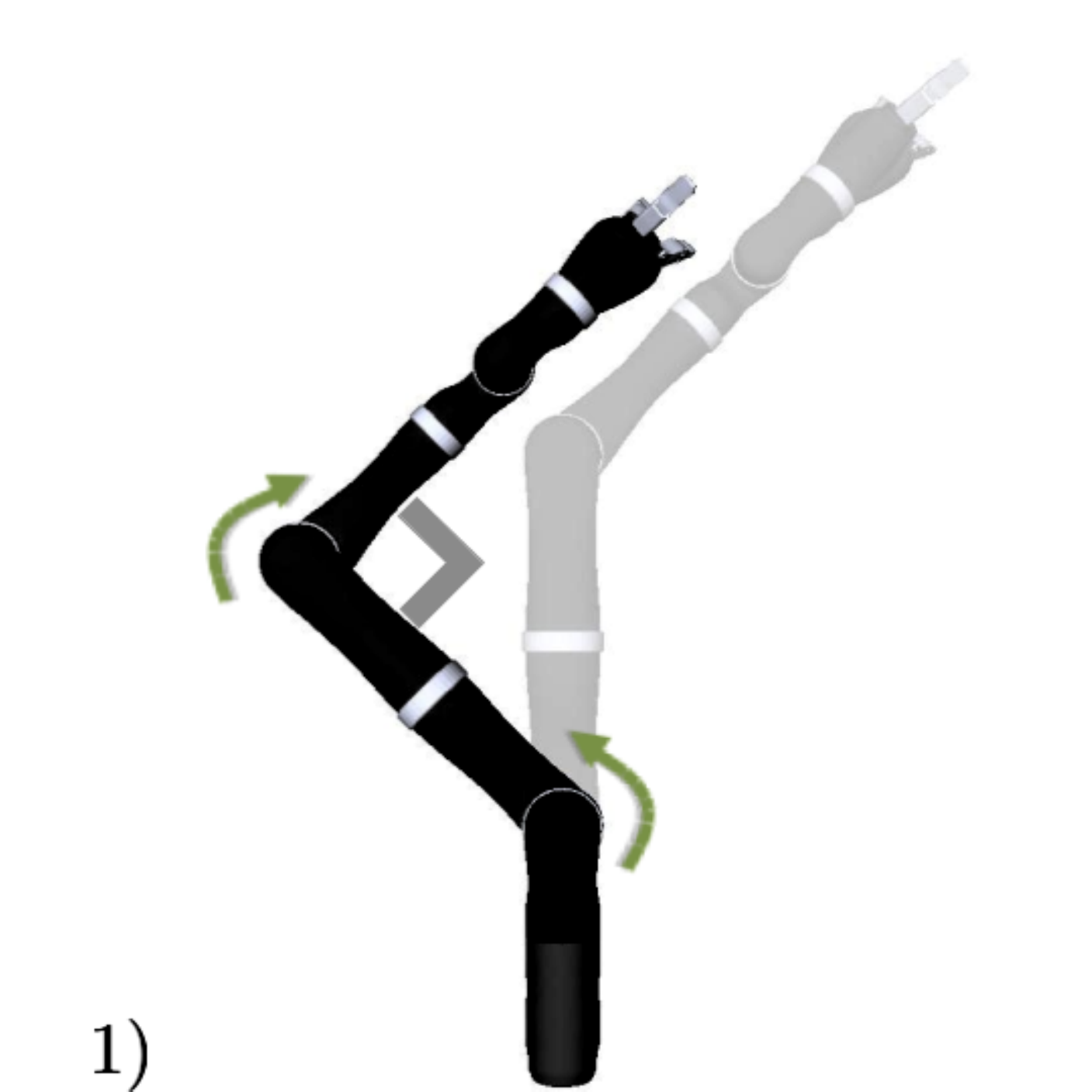}  & \includegraphics[height=1.in, width=1.in]{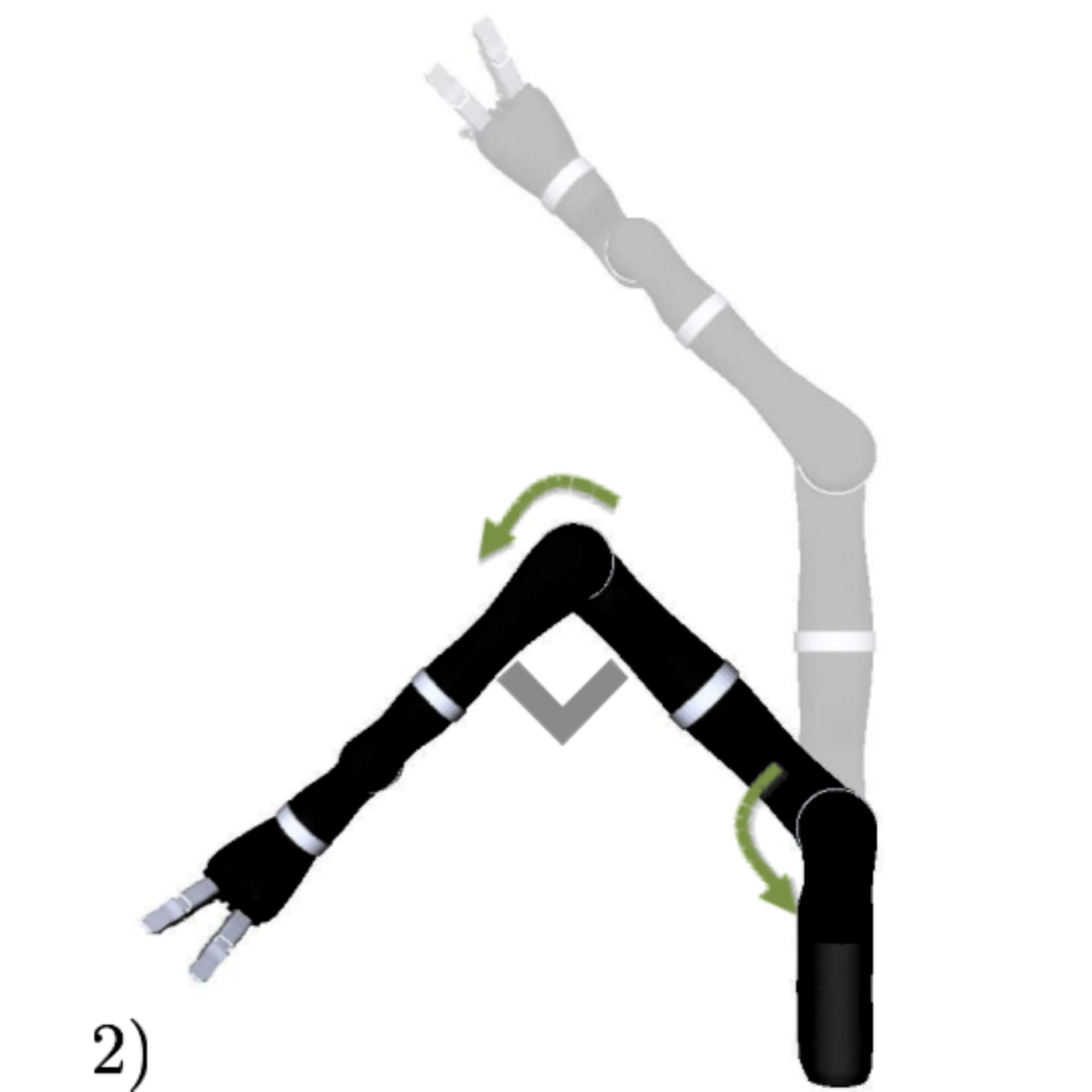} \\
        \includegraphics[height=1.in, width=1.in]{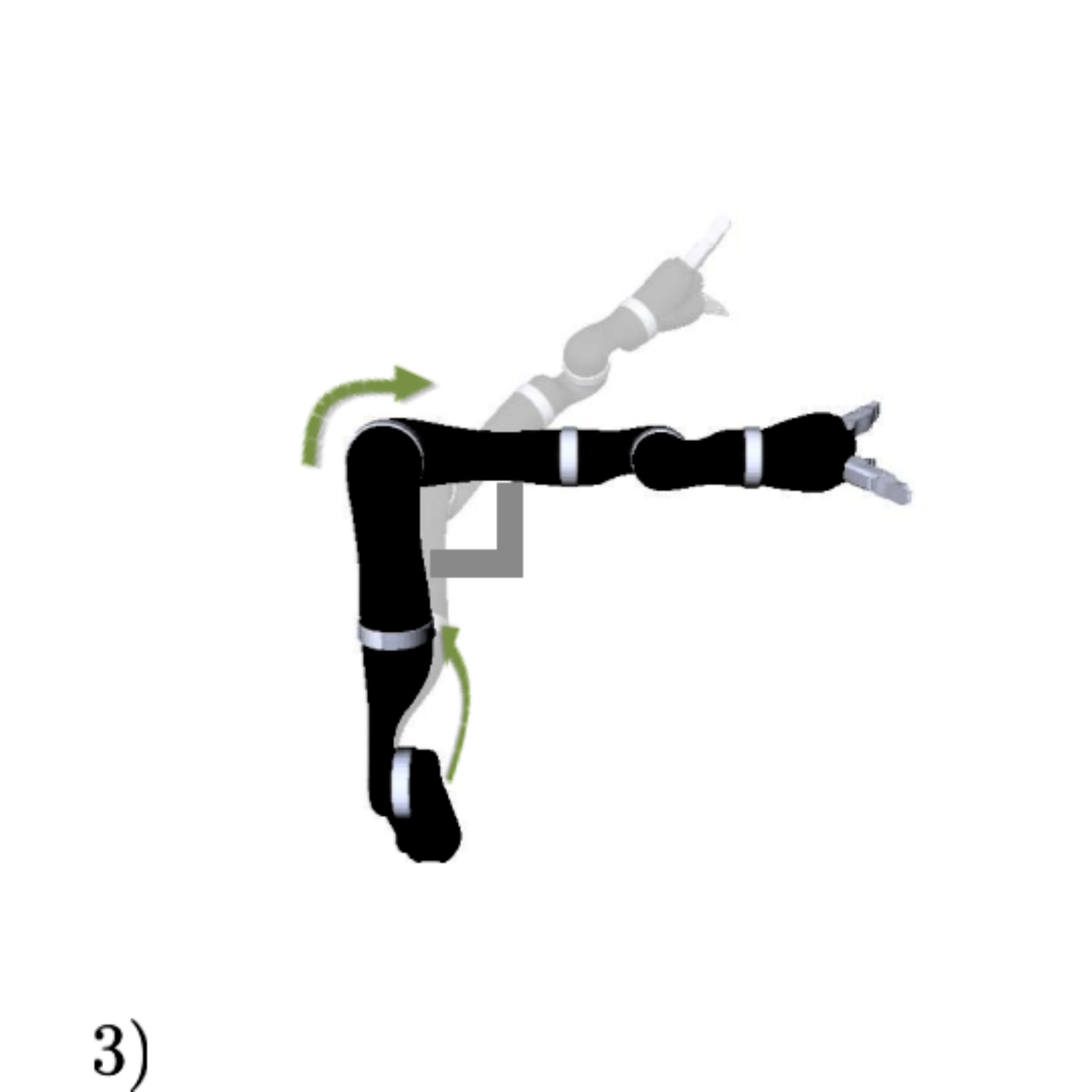} & \includegraphics[height=1.in, width=1.in]{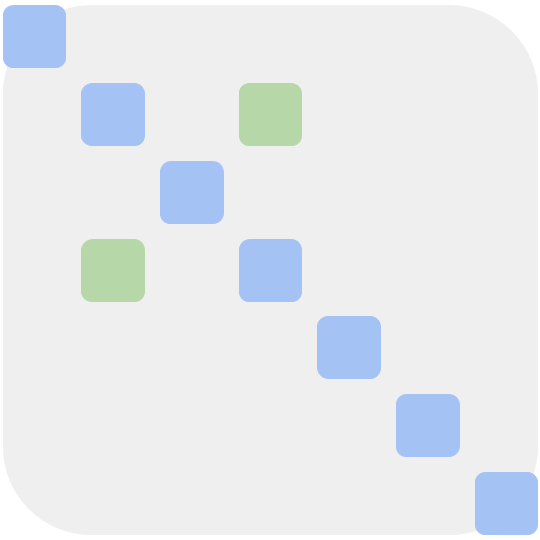} \\
    \end{tabular}
    \caption{Results of satisfying an elbow $90^{\circ}$ constraint on 3 different starting configurations, with a metric that positively correlates the shoulder and elbow joints. While 1) and 2) are mirror images of each other, their solutions are not. In 3), the shoulder actuates, as the elbow moves to $90^{\circ}$, moving the arm out of the page. With this fixed positive correlation, the robot's behavior does not generalize well across different starting configurations.}\label{fig:constantproblems}
    \vspace{-0.2cm}
\end{figure}

\subsection{ Metrics and Correlations}
While Gielnaik et al. \cite{gielniak_thomaz_2011, todorov_jordan_2002}, posit that spatially correlating the joints of a robot will lead to more human-like robot motion, we find that for robot arms (as opposed to humanoids), this spatial correlation isn't robust across different starting configurations.

When we examine correlation in 7 dimensional C-Space
, we find that joint correlations can not be agnostic to the start configuration. \figref{fig:constantproblems} demonstrates problems that arise from \emph{\textbf{positive}} correlation between joints $1$ (shoulder pitch) and $3$ (elbow). From the starting configuration (transparent) in \figref{fig:constantproblems} (1), with shoulder yaw $=0$, actuating the elbow moves the shoulder in a direction that looks natural. However, if we apply this same metric and to the starting configuration in \figref{fig:constantproblems} (2) with shoulder yaw =$\pi$, the orientations of the joints becomes reversed. Decreasing joint $3$'s value moves the forearm counterclockwise now instead of clockwise while joint $1$ retains is orientation. Visually, \figref{fig:constantproblems} (2) is the mirror image of \figref{fig:constantproblems} (1) so we would expect it to produce a mirrored end configuration but this is not the case. The orientation flip now requires a \emph{\textbf{negative}} correlation to produce the natural mirrored solution.

Worse yet, \textit{adversarial} starting configurations like \figref{fig:constantproblems} (3) with shoulder yaw $=\frac{\pi}{2}$ can produce even more undesired joint coupling. Here, the rotational axis of joint $1$ is orthogonal to that of joint $3$. Moving joint $3$ will also move joint $1$ from correlation but the end result is nothing like the natural result of \figref{fig:constantproblems} (1). We wanted positive correlation in \figref{fig:constantproblems} because it produced a natural robot configuration but it came at the cost of messing up behavior in different starting configurations. We cannot rely on fixed \textit{joint correlation} terms in 7D space to be robust across all starting configurations. 

\subsection{Learned Metrics}
We learn a diagonal metric in 7 dimensions
using the same learning algorithm employed in the 3 dimensional case. We ignore correlations in this analysis because of our finding from above -- even so, the analysis should tell us whether Euclidean joint costs are appropriate.

We again learn separate metrics for contraction and expansion tasks but now we dress up the end effector location constraint in a real-world scenario. The contraction task is disguised as the robot bringing a book from a bookshelf closer to the robot base. The expansion task is shown in \figref{fig:Learned7DOF}: the robot reaching for a book in the bookshelf.

To collect data, we queried 20 participants (mean age of 33, female $30\%$) via Amazon Mechanical Turk. All participants were from the United States and had a minimal approval rating of $95\%$. Each subject answer 36 questions (18 contractions, 18 expansions). We used the same 4 criteria from the 3DOF experiments.

\noindent\textit{Euclidean Better for Contraction.}
From \figref{tab:kl_7dof}, we notice that in 7DOFs, \emph{\textbf{contractions}} perform significantly better with the Euclidean metric than \emph{\textbf{expansion}}. This is contrary to what occurred with 3DOFs.

\begin{table}
    \vspace{0.45cm}
    \caption{KL Divergence for Euclidean and Learned Metric}
    \label{tab:freq}
    \centering
    \begin{tabular}{ccc}
    \toprule
    & Euclidean & Learned\\
    \toprule
    Naturalness Contraction & 5.0758441747 & 3.5453962573 \\
    Naturalness Expansion & 11.26325615 & 2.2152094803 \\
    \midrule
    Similarity Contraction & 4.7339982443 & 2.6521939985 \\
    Similarity Expansion & 12.05742235 & 2.7618380878 \\
    \midrule
    Closeness Contraction & 5.6072539842 & 2.790643385 \\
    Closeness Expansion & 11.556934461 & 2.4291169369 \\
    \midrule
    Predictability Contraction & 4.8851149143 & 2.92711599519 \\
    Predictability Expansion & 11.498342560 & 3.2229499891 \\
  \bottomrule
\end{tabular}\label{tab:kl_7dof}
\vspace{-0.4cm}
\end{table}

\noindent\textit{Learned Metrics for Expansion.}
The learned expansion metric for all 4 criteria included an expensive joint $3$ (elbow). It is interesting that across both the 3DOF and 7DOF cases, expansion metrics consistently prefer an expensive elbow joint. Additionally, shoulder roll, shoulder yaw, and wrist roll were all very cheap. This allowed the learned metric to perform the wrist flip in \figref{fig:Learned7DOF}.
expensive elbow again

\noindent\textit{Learned Metrics for Contraction.}
For contractions, the learned metrics consistently had expensive shoulder pitch and yaw. This is the intuitive result because motion in the shoulder moves the entire arm more than motion in other joints (i.e. elbow or wrist). This would lead to robot configurations that users would find less visually similar and further away from the start configuration.

\section{Summary of Findings}
Overall, contraction and expansion tasks tend to determine how good of a fit the Euclidean metric is. For 3DOF arms, expansion tasks are well fit by the Euclidean metric. Contraction tasks in 3DOFs require tuning to match human preferences, specifically expensive elbow with strong positive shoulder-wrist correlation. In 7DOFs, we neglected correlations for better robustness and found that the Euclidean metric performed well on contraction tasks. After learning a metric from 7DOF contraction tasks, we recovered the intuitive expensive shoulder metric. 7DOF Expansion tasks needed learning and resulted in an expensive elbow joint, like for 3DOFs. Lastly, across these robots and tasks, we consistently saw expensive elbow cost. From all this, we have reason to believe that for robots to act naturally and predictably, their notion of distance should be more sophisticated on some tasks than the Euclidean metric in C-Space. In the future, we hope to conduct experiments demonstrating the merits of different metrics when integrated in various motion planning algorithms i.e. trajopt and RRT \cite{Lavalle98rapidly-exploringrandom}.


\section*{Acknowledgments}
This research was supported by funding from the AFOSR and NSF CAREER Award.
\bibliographystyle{plain}
\bibliography{biblio}

\end{document}